\documentclass[10pt,twocolumn,letterpaper]{article}

 \usepackage{cvpr}              
\usepackage{float}
\usepackage{graphicx}
\usepackage{afterpage}
\usepackage[T1]{fontenc}
\usepackage{amssymb}
\usepackage{pifont}
\usepackage{multirow}
\usepackage{caption}
\usepackage{xcolor}
\usepackage{siunitx}
\usepackage{comment}
\usepackage{tabularx}
\usepackage[accsupp]{axessibility} 
\definecolor{cvprblue}{rgb}{0.21,0.49,0.74}
\usepackage[pagebackref,breaklinks,colorlinks,allcolors=cvprblue]{hyperref}

\def\paperID{} 
\def\confName{CVPR}
\def\confYear{2026}

\title{TacSIm: A Dataset and Benchmark for Football Tactical Style Imitation}

\author{Peng Wen \quad Yuting Wang \quad Qiurui Wang \thanks{Corresponding author}\\
}
\begin{document}
\twocolumn[{
	\maketitle 
	\renewcommand{\twocolumn}[1][]{#1}
	\begin{center}
		\captionsetup{type=figure}
		\includegraphics[width=\textwidth]{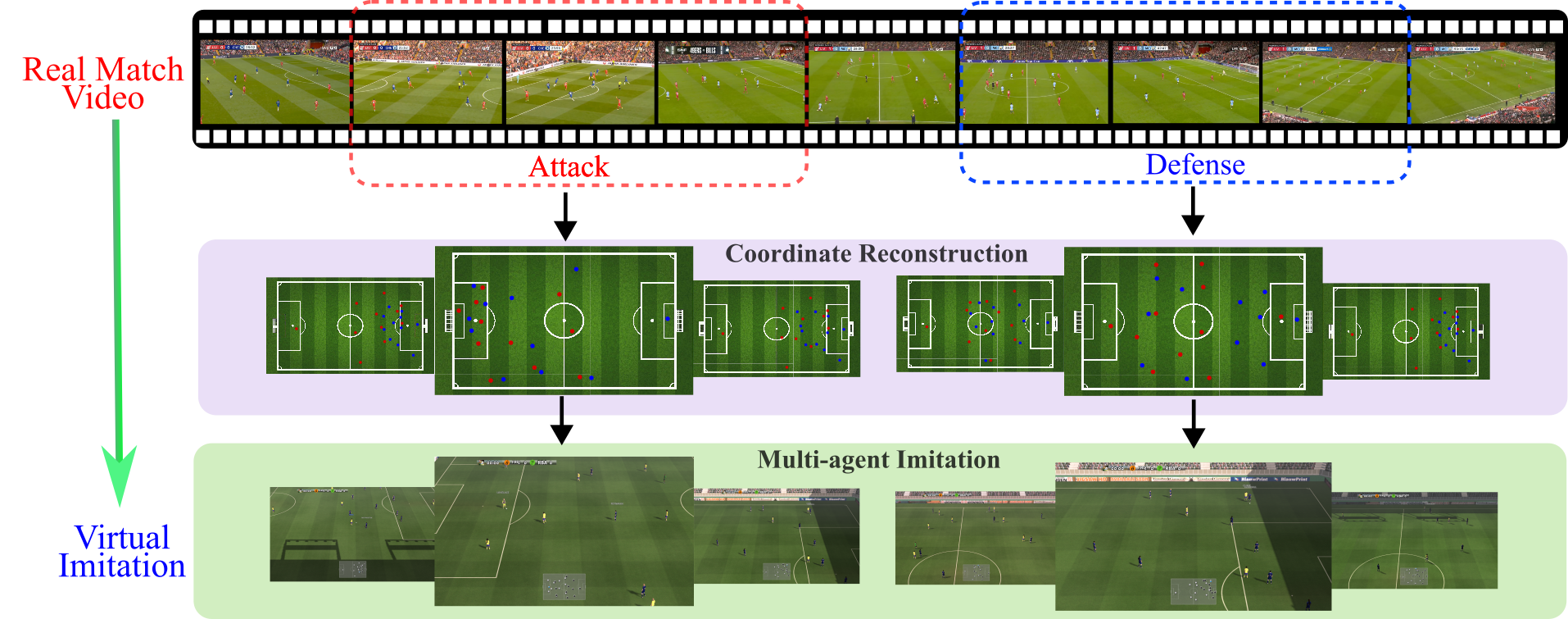}
		\captionof{figure}{Overview of \textit{TacSIm}. (a) Video matches in real world: Frames captured from televised football matches, segmented into offensive and defensive phases to display players' real-time positions on the field. (b) Player coordinate reconstruction: Mapping real-time player positions from broadcast frames to a normalized pitch coordinate system. (c) Tactical replication in a virtual football environment. 
		The current contexts of reconstructed players, such as the actions and positions, are fed into a virtual football simulation platform where each player is treated as an agent. by multi-agent system learning, the following contexts of each player are reproduced and can be compared with the real contexts of each player in the following time of the real football match. The reproduced behaviors in virtual football environment can be visualized.
		}
	\end{center}
}]
\footnotetext[1]{Corresponding Author.}
\begin{abstract}
	Current football imitation research primarily aims to optimize reward-based objectives, such as goals scored or win rate proxies, paying less attention to accurately replicating real-world team tactical behaviors. We introduce TacSIm, a large-scale dataset and benchmark for \textbf{Tac}tical \textbf{S}tyle \textbf{Im}itation in football. TacSIm imitates the acitons of all 11 players in one team in the given broadcast footage of Premier League matches under a single broadcast view. 
	Under a offensive or defensive broadcast footage, TacSIm projects the beginning positions and actions of all 22 players from both sides onto a standard pitch coordinate system. 
	TacSIm offers an explicit style imitation task and evaluation protocols. Tactics style imitation is measured by using spatial occupancy similarity and movement vector similarity in defined time, supporting the evaluation of spatial and temporal similarities for one team. 
	We run multiple baseline methods in a unified virtual environment to generate full-team behaviors, enabling both quantitative and visual assessment of tactical coordination. By using unified data and metrics from broadcast to simulation, TacSIm establishes a rigorous benchmark for measuring and modeling style-aligned tactical imitation task in football.
	The dataset and benchmark are available at \href{https://github.com/wenpengpap/TacSIm}{TacSIm}.
\end{abstract}  
\section{Introduction}
\label{sec:intro}
Football tactical imitation refers to learning and replicating a team's spatio-temporal organization patterns within real match scenarios, encompassing elements such as formation and positioning, tempo control and off-the-ball coordination. It focuses on intricate tactical structures rather than merely aggregate metrics like win rates~\cite{bauer2021automated}. For football teams, the importance of tactical imitation lies in preserving and executing the stylistic characteristics of the team, transforming the abstract principles of “how we play” into replicable measurable behavioral representations~\cite{teixeira2025mapping}. 
It can serve as scenario testing for coaches and customized opponent planning within simulation environments. It also supports player development and recruitment decisions through role adaptation and tactical response evaluations while providing standardized interpretable objectives for analysis beyond raw match outcomes~\cite{shao2024virtual}. By aligning data-driven models with real tactical structures, tactical mimicry effectively bridges the gap between simulation and reality~\cite{pu2024orientation,cruz2021bridging}. Tactical imitation improves generalization capabilities in different opponents and match phases. Furthermore, it provides a safer data-driven sandbox for exploring ``what-if'' strategies before game time.

Although conceptually compelling, tactical imitation remains challenging in replicating a team's overall style in practice~\cite{teixeira2025mapping}. Firstly, data acquisition is constrained. Many researches rely on synthetic or limited tracking data, making it difficult to obtain 11-v-11 full-team trajectories sourced from broadcast footage.
Raw tracking and event-level data from top leagues remain commercially restricted and inaccessible. Broadcast-based data are vulnerable to multi-camera switching, camera motions (such as zoom/pan/tilt), visual obstructions from crowds and field markings, inconsistent frame rates and resolutions~\cite{davis2024methodology}. Secondly, in imitation process, approaches often exhibit imbalances between prioritizing individual-level behavior cloning and pursuing reward-maximizing agent learning.
Under partially observable conditions, such methods also demonstrate relatively weak generalization capabilities regarding opponents, match tempo and game phases~\cite{corsie2021spatial,russo2025leveraging}. Thirdly, evaluation frameworks predominantly emphasize individual errors or segment-level rewards rather than assessing the spatial and temporal consistency of actions between real team and model-generated  results~\cite{wagenmaker2024overcoming}.

To bridge this gap, we introduce TacSIm, a new dataset and benchmark for multi-agent tactical imitation in football. TacSIm reconstructs trajectories of all players and the ball from the given broadcast footage, capturing offensive and defensive phases across diverse match contexts. Unlike existing datasets emphasizing individual actions or short-term events, TacSIm enables learning and evaluation at the tactical level. Researchers can deploy models trained on real trajectories within virtual football environments to test their ability to reproduce coordinated team behaviors.

Based on this dataset, we establish the first tactical imitation benchmark, whose core task is to reproduce team-level strategies from real match. Furthermore, we compare multi-agent tactical imitation approaches by uniformly modeling the spatio-temporal dynamics of each agent and learning coordinated decision strategies. To ensure comparability among methods and facilitate future research, we provide standardized evaluation protocols and metrics to quantify team coordination and tactical reproduction.

Our contributions are:
\begin{itemize}
	\item[$\bullet$]To our knowledge, we are the first to propose a tactical imitation dataset derived from broadcast footage and camera calibration. By reconstructing trajectories and performing coordinate transformations on monocular broadcast footage, we uniformly project player and ball positions onto a standard field coordinate system, covering top-tier match scenarios across multiple teams and different game phases.
	\item[$\bullet$]We define a standardized football tactics imitation task and evaluation protocols, quantifying virtual-reality consistency through space-based, time-dependent metrics. 
	We simultaneously report two metrics: (I) spatial occupancy overlap and (II) consistency in movement direction patterns. We aggregate them into a composite score which can reflect tactical style fidelity.
	\item[$\bullet$]Under a unified evaluation protocol, typical existing models are used to quantify the ability of football tactical style imitation to reveal the advantages and disadvantages of them.
\end{itemize}
We hope TacSIm will serve as a catalyst for bridging computer vision and multi-agent learning in sports analytics, paving the way toward a unified understanding of collective intelligence in both real and virtual football.

\section{Related work}
\label{sec:formatting}
\subsection{Analytics Datasets in Football}
Tactics research in football benefits from an expanding ecosystem of datasets capturing visual, event-based and trajectory-level representations of matches. Early datasets, such as SoccerNet~\cite{giancola2018soccernet,deliege2021soccernet}, focus on large-scale video understanding, providing annotations for goals, passes and other broadcast-level events. Although these resources promote progress in temporal action detection and player identification, they lack fine-grained spatio-temporal data that can describe full-team coordination.

Event-based datasets extend research by incorporating structured annotations of discrete-time passing, shooting and positional data, enabling the calculation of tactical statistics and the prediction of outcomes ~\cite{pappalardo2019public,pappalardo2019playerank,umemoto2023evaluation}. However, their low temporal resolution and lack of continuous trajectories render them unsuitable for modeling multi-agent cooperative behavior. Synthetic environments, such as Google Research Football (GRF)~\cite{kurach2020google}, provide fully observable and controllable match imitations that expose the state of all agents and the ball at every time step. Although these simulated datasets are invaluable for reinforcement learning and policy optimization, they exhibit significant differences from real-world broadcast dynamics and lack consistency with authentic tactical scenarios.

\subsection{Imitation Learning}
Imitation learning offers a data-driven framework for replicating expert behaviors from demonstrations and has become a crucial approach in autonomous driving~\cite{pan2020imitation}, robotics~\cite{celemin2022interactive} and multi-agent learning~\cite{wu2024cbil,Liu2020Multi-Agent}. 
Classical lines include Behavior Cloning(BC)~\cite{pomerleau1988alvinn} and the family of Inverse Reinforcement Learning(IRL)~\cite{rahimian2021inferring,adams2022survey}, which first infers a reward function that explains expert demonstrations and then optimizes a policy under that reward; representative variants include maximum-entropy IRL~\cite{song2025survey} and adversarial IRL (AIRL)~\cite{yu2019multi}. 
Adversarial imitation such as GAIL~\cite{ho2016generative} can be viewed as learning a surrogate reward via a discriminator. 
Recent extensions target hierarchical structures and multi-agent coordination, enabling policies that model inter-player dependencies~\cite{mu2025hierarchical,yu2022surprising,nicholaus2025ftpsg}.

In sports, imitation learning has been applied to model decision-making and motion coordination, often utilizing reinforcement learning backbones or transformer-based architectures to capture long-term dependencies between agents~\cite{le2017coordinated,Liu2020Multi-Agent,wang2024tacticai}. However, most of these approaches rely on synthetic or proprietary datasets, which limit tactical realism and diversity.

\section{Dataset Construction}
\subsection{Data Acquisition}
\begin{table*}[htbp]
	\centering
	\caption{
		\textbf{Dataset comparison} showing the scope, scale and tactical capability.
		TacSIm uniquely provides real broadcast trajectories with tactical semantics and supports real-to-virtual simulation.
	}
	\label{tab:dataset_comparison}
	\renewcommand{\arraystretch}{1.2}
	\setlength{\tabcolsep}{1pt}
	\small
	\begin{tabular}{lcccccc}
		\toprule
		\multirow{2}{*}{\textbf{Dataset}} &
		\multirow{2}{*}{\textbf{Task}} &
		\multicolumn{2}{c}{\textbf{Data Content}} &
		\multicolumn{2}{c}{\textbf{Tactical Attributes}} &
		\multirow{2}{*}{\textbf{Availability}} \\
		\cmidrule(lr){3-4} 
		\cmidrule(lr){5-6} 
		&  & Player Trajectories & Ball Position & Tactical Phases & Simulation Capability &  \\
		\midrule
		SoccerNet-v2~\cite{deliege2021soccernet} & Event detection & \ding{55} & \ding{55} & 
		\ding{55} & \ding{55} & Public \\
		SoccerNet-Tracking~\cite{cioppa2022soccernet} & Multi-object tracking & \ding{51} & \ding{51} & 
		\ding{55} & \ding{55} & Public \\
		StatsBomb 360~\cite{umemoto2023evaluation} & Context analytics & \ding{55} & \ding{51} & 
		\ding{55} & \ding{55} & Licensed \\
		Metrica Sports~\cite{kim2023ball} & Trajectory visualization & \ding{51} & \ding{51} & 
		\ding{55} & \ding{55} & Public \\
		\midrule
		\textbf{TacSIm (Ours)} & Tactical imitation & \textbf{\ding{51}} & \textbf{\ding{51}} & 
		\ding{51} (attack / defense) & \ding{51} (real-to-virtual) & Public \\
		\bottomrule
	\end{tabular}
\end{table*}
The raw material for TacSIm originates from official broadcast footage of the 2024-2025 English Premier League (EPL) season, covering 140 matches across all EPL teams. The data are selected for its tactical diversity, consistent participation in high-level matches and the superior quality of the broadcast footage. All videos are captured at 1080p resolution and 25 FPS and preprocessed to retain only in-play sequences, removing replays, halftime breaks, advertisements and non-game camera shots to preserve temporal and tactical continuity.

\textbf{Clip Selection Strategy.} To focus on collective tactical behaviors rather than isolated individual actions, match footage is manually segmented into possession-based clips. Each segment corresponds to a complete ball-possession cycle, defined as the period from when a team gains control of the ball until they lose it, thereby ensuring that each clip captures a coherent offensive or defensive process. A clip is included if it (a) covers the entire possession phase without interruptions, such as stoppages and referee interventions, (b) excludes excessive camera cuts, replay sequences, or off-play scenes that may disrupt temporal continuity, and (c) represents a valid and complete possession phase that encompasses the full range of attacking and defensive sequences observed in real match conditions, without being restricted to a predefined set of tactical patterns. Initial segmentation is performed by the research team and subsequently refined through detailed review and annotation by trained domain experts to ensure temporal accuracy and contextual validity.

\textbf{Annotation Scheme.} Following the official annotation guideline, we adopt a phase-level hierarchical labeling taxonomy to describe the tactical organization of football matches. Each possession segment is categorized as Attack, Defense, or Transition, determined primarily by the ball possession status.  The Attack and Defense phases are further subdivided into outcome-based subtypes to capture tactical intent and result. A \textbf{Successful Attack} refers to an offensive phase that concludes with a shot or a clear scoring attempt, whereas a \textbf{Failed Attack} terminates with ball loss before a shot occurs.  Conversely, a \textbf{Successful Defense} represents a defensive phase in which the defending team regains possession or prevents a shot from being taken. At the same time, a \textbf{Failed Defense} denotes a sequence that allows a shot or concedes a set piece.  Transition phases are used for ambiguous or rapidly changing moments, such as counter-pressing and contested recoveries, where possession status shifts within a short temporal window. The annotation guideline provides explicit decision-flow charts and rule-based criteria to resolve ambiguous cases. It ensures high inter-annotator consistency and reproducibility across the dataset.

\textbf{Annotator Qualification.} All annotations in TacSIm are performed by a team of annotators with formal academic training in football tactics and performance analysis. The annotators consist of undergraduate and graduate students majoring in football science at sports universities, each possessing substantial knowledge of tactical principles, match organization and positional structures. Their background coursework includes modules in match strategy, game phase analysis and data-driven performance evaluation, ensuring that they could accurately interpret complex tactical patterns rather than rely on surface-level visual cues.

\subsection{Data Processing and Tactical Reconstruction}
After collecting and annotating the broadcast footage, we process all videos to obtain precise two-dimensional trajectories of every player and the ball, forming the foundation for tactical replay and imitation tasks. Following the SoccerNet-GSR paradigm~\cite{somers2024soccernet}, our pipeline has two stages: (1) \textit{In-camera position reconstruction}, where detections are tracked and projected to standardized pitch coordinates; and (2) \textit{Off-camera position reconstruction}, where missing states are imputed using a conditional VAE~\cite{qi2020imitative}. 

\textbf{In-camera position reconstruction}. We follow the SoccerNet-GSR~\cite{somers2024soccernet} formulation, adapting it to the camera configuration of our dataset. 
Each match video is decomposed into frame sequences and a YOLOv11-based detector identifies all visible players and the ball in each frame. 
Detections are associated temporally via a DeepSORT~\cite{mansourian2023multi} tracker that maintains consistent player identities throughout occlusions. 
For each frame, we use TVCalib~\cite{theiner2023tvcalib} estimate camera parameters from detected pitch keypoints (sidelines, penalty boxes, center circle) to compute the homography for transforming image coordinates into a standardized bird's-eye-view pitch space. Player positions are projected by treating the bottom center of each bounding box as the ground-contact point. 
When calibration lines are missing, homographies are interpolated between adjacent frames or recovered using robust solvers. 
All coordinates are projected into a coordinate system with $x \in [-1, 1]$ and $y \in [-0.42, 0.42]$ for simulation in the GRF~\cite{kurach2020google} environment, as this matches the field dimensions of the GRF setup.

\textbf{Off-camera position reconstruction}.
Broadcast data often suffer from occlusions, rapid camera transitions or incomplete detections, resulting in fragmented trajectories. The VAE framework demonstrates significant advantages in reconstructing continuous, physically plausible trajectories. By learning the probability distribution in the latent space, it not only generates smooth and diverse motion sequences but also effectively captures trajectory uncertainty~\cite{ivanovic2020multimodal,omidshafiei2021time,wu2024cbil,kingma2013auto}. We adopt the trajectory reconstruction reconstruction method~\cite{qi2020imitative}. Given a partially observed sequence \( X^{\text{obs}} \in \mathbb{R}^{N \times T \times 2} \) and observation mask \( M \), our model learns \( p_\theta(X^{\text{full}}\mid X^{\text{obs}},M) \) via a latent variable \( z \sim p(z) \) and approximate posterior \( q_\phi(z\mid X^{\text{obs}},M) \). The training objective minimizes the reconstruction and regularization losses:
\begin{equation}
	\begin{aligned}
		\mathcal{L} = \mathbb{E}_{q_\phi(z \mid X^{\text{obs}}, M)} \big[ \|(1-M) \odot (X - \hat{X}_\theta(z, X^{\text{obs}}, \\ M)) \|_2^2 \big]
		+ \beta \cdot \mathrm{KL}\big[ q_\phi(z \mid X^{\text{obs}}, M) \parallel p(z) \big].
	\end{aligned}
\end{equation}
The encoder utilizes a Bi-directional RNN to encode individual motion and inter-agent dependencies while the decoder reconstructs complete trajectories conditioned on the latent variable \(z\). Additional regularizers enforce motion smoothness, formation coherence and collision avoidance. Short-term gaps (\(<5\) frames) are interpolated by using motion-constrained splines, whereas the VAE imputer fills longer missing segments. Training is conducted with random masking, temporal augmentations and velocity-based noise to improve generalization. We train on the soccer trajectory Alfheim Dataset~\cite{pettersen2014soccer}. See the Supplementary material.
\begin{figure}[t]
	\centering
	\includegraphics[width=\linewidth]{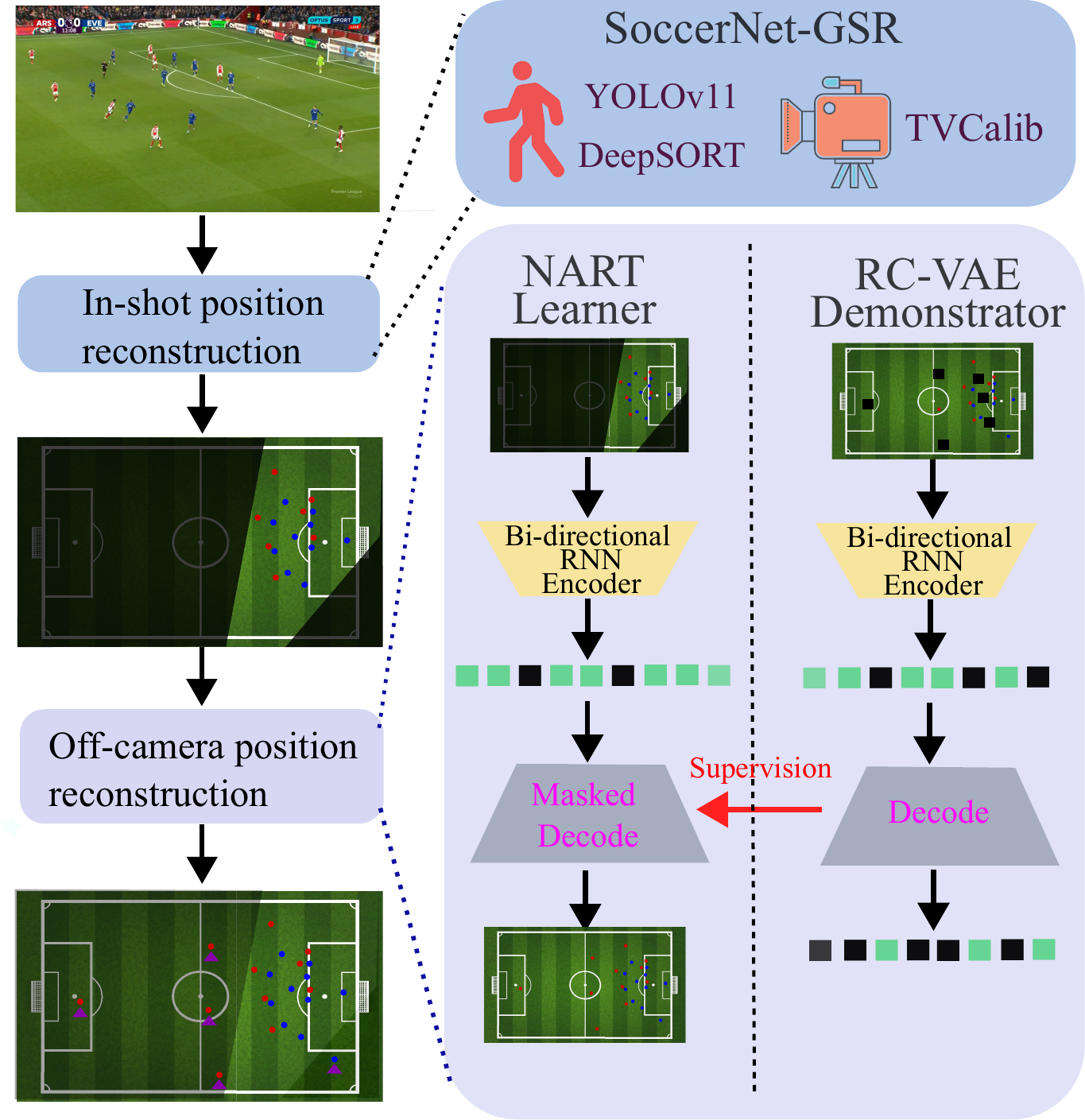}
	\caption{
		Overview of the trajectory completion framework with demonstrator–learner imitation structure.}
	\label{fig:onecol}
\end{figure}

As shown in Figure \ref{fig:onecol}, the system first extracts players and ball coordinates from broadcast videos using the SoccerNet-GSR method, projecting them into a bird’s-eye view. The Demonstrator (right) is a Bi-directional RNN-based non-autoregressive model that observes complete trajectories and learns spatiotemporal dynamics by encoding and decoding the full sequence. The Learner (left) receives masked or partially observed trajectories as input and employs a masked decoder to reconstruct the missing motion. During training, the learner’s decoder is supervised by the demonstrator’s decoded representations, allowing for the imitation of the demonstrator’s temporal dependencies and motion patterns. The completed trajectories are then projected back to the field for evaluation and visualization.

\textbf{Data Split.} 
All experiments are conducted on the proposed TacSIm dataset, which comprises 194,565 annotated video segments spanning approximately 38,913 seconds of Premier League broadcast footage. 
To ensure fair evaluation and prevent team-specific data leakage, we perform the split strictly by match identity, assigning 70\% of matches to the training set, 15\% to validation, and 15\% to testing. 
This split preserves the distribution of tactical phases and ensures comparable coverage of both offensive and defensive contexts. 
Table~\ref{tab:dataset_split} provides detailed statistics of the dataset composition.
\begin{table}[t]
	\centering
	\caption{
		Dataset split statistics for the TacSIm.
		We report the number of video segments, total duration, and proportion for each split.
		All data are sampled from 180 hours of Premier League broadcasts with balanced coverage of offensive and defensive phases.
	}
	\label{tab:dataset_split}
	\small  
	\setlength{\tabcolsep}{2pt}
	\renewcommand{\arraystretch}{1.4} 
	\begin{tabular*}{\linewidth}{@{\extracolsep{\fill}}lccc}
		\toprule
		\textbf{Split} & \textbf{Total Duration (s)} & \textbf{Video Segments} & \textbf{Proportion (\%)} \\
		\midrule
		Training   & 136,195 & 27,240 & 70 \\
		Validation & 29,185  & 5,837  & 15 \\
		Test       & 29,185  & 5,836  & 15 \\
		\midrule
		\textbf{Total} & \textbf{194,565} & \textbf{38,913} & \textbf{100} \\
		\bottomrule
	\end{tabular*}
\end{table}

\textbf{Integration and Replay.} After imputation and final kinematic filtering, each possession segment is represented as a continuous sequence of player and ball coordinates $\{x_{i,t},y_{i,t}\}_{i=1}^{22}$ and $(x_{ball,t},y_{ball,t})$, synchronized with the corresponding tactical phase label. 
These trajectories enable realistic tactical reconstruction, visualization and multi-agent imitation benchmarking on virtual football platforms.

\section{Benchmark Design}
To evaluate the ability of models to reproduce realistic team-level tactics, TacSIm defines a spatial-temporal benchmark for tactical imitation. The benchmark measures how closely a model-generated trajectory sequence aligns with the corresponding ground-truth sequence extracted from real matches, both in terms of spatial occupation and directional movement coherence.

Spatial Discretization. The football pitch is discretized into a grid of uniform cells to convert continuous trajectories into spatial occupancy representations. Each player’s ground-plane coordinates are mapped onto this grid over time, producing a binary occupancy tensor \( \mathbf{O}_{i,t} \in \{0,1\}^{H \times W} \), where \( H \) and \( W \) denote the grid resolution. Five distinct grid sizes are chosen to capture a wide range of tactical behaviors in football. Coarse grids (low resolution) are used to track fast-paced plays such as counter-attacks, where the emphasis is on overall movement flow rather than fine-grained positioning. Medium grids (mid-resolution) are used for general offensive or defensive phases that require a balance between flow and positional accuracy. Fine grids (high resolution) are used for slower buildup plays and set-pieces, where precise player positioning and formation are critical. Very fine grids are applied to analyze specific areas of the pitch, such as the penalty box or midfield, during key moments that demand attention to small player movements. Extra-fine grids are used to capture highly detailed tactical scenarios, such as intricate dribbling or precise passing sequences. These grid sizes are chosen to strike a balance between spatial precision and computational efficiency, enabling detailed analysis of various types of gameplay, from high-speed transitions to intricate tactical setups.

Let $s_t$ denote the average displacement magnitude within a temporal window; the grid size $\Delta_g$ is dynamically scaled as
\begin{equation}
	\Delta_g = \min(\Delta_{\text{max}}, \max(\Delta_{\text{min}}, \alpha / s_t)).
\end{equation}
where $\alpha$ is a tunable scaling coefficient and $\Delta_{\text{min}}, \Delta_{\text{max}}$ denote the lower and upper grid limits. 
This adaptive discretization ensures that spatial comparisons remain consistent across tactical contexts with different motion intensities.

\textbf{Imitation Metrics}. We compute two complementary similarity measures:

\textbf{(1) Spatial Occupancy Similarity.}
Given the binary occupancy maps of the ground-truth sequence $\mathbf{O}^{\text{gt}}$ and the model-generated sequence $\mathbf{O}^{\text{pred}}$, to quantify spatial overlap between real and predicted occupancies, we employ the Jaccard index to measure spatial occupancy similarity:
\begin{equation}
	S_{\mathrm{t}}
	= \frac{\bigl|\mathbf{O}^{\mathrm{gt}} \cap \mathbf{O}^{\mathrm{pred}}\bigr|}
	{\bigl|\mathbf{O}^{\mathrm{gt}} \cup \mathbf{O}^{\mathrm{pred}}\bigr|}
	\in [0,1].
\end{equation}
The metric reflects the proportion of shared occupied cells relative to the union of both sets. Higher values indicate stronger agreement in spatial coverage and team formation.

\textbf{(2) Movement Vector Similarity.}
To assess directional and activity alignment, we flatten the occupancy tensors into motion feature vectors $\mathbf{v}^{\text{gt}}$ and $\mathbf{v}^{\text{pred}}$, encoding both cell activation and temporal frequency. 
Cosine-like similarity is then computed as
\begin{equation}
	S_{\mathrm{v}} 
	= \frac{1}{2}\!\left(\frac{\mathbf{v}^{\mathrm{gt}}\!\cdot\!\mathbf{v}^{\mathrm{pred}}}{\|\mathbf{v}^{\mathrm{gt}}\|\,\|\mathbf{v}^{\mathrm{pred}}\|} + 1\right) \in [0,1].
\end{equation}
When either vector has zero norm, we set $S_{\mathrm{v}}{=}1$ if both of the vectors are zero (complete agreement in inactivity) and $S_{\mathrm{v}}{=}0$ otherwise. 
While the Spatial Occupancy Similarity emphasizes positional alignment, the Movement Vector Similarity captures flow-level agreement in movement direction and activity patterns between predicted and real trajectories.

\textbf{Evaluation Protocol.} For each possession segment, the model is required to generate a sequence of multi-agent trajectories conditioned on the observed context. 
The similarity scores $S_{\text{t}}$ and $S_{\text{v}}$ are computed for all agents within the segment and averaged over time and across the dataset. 
The final benchmark score is defined as the harmonic mean of the two similarity measures:
\begin{equation}
	\text{Score} = \frac{1}{2}({S_{\text{t}} + S_{\text{v}}}).
\end{equation}
This combined index jointly reflects spatial occupation accuracy and dynamic consistency, providing a holistic measure of tactical imitation fidelity. The use of the harmonic mean is particularly well-suited for this task since it emphasizes lower values when there is a significant disparity between the two similarity scores. This ensures that both spatial consistency and dynamic consistency are treated with equal importance. The model’s performance is penalized if either measure is significantly worse. The harmonic mean prevents overestimating performance when one metric is much higher than the other, making the final score a more balanced and fair representation of the model’s ability to mimic realistic team-level tactics.

\textbf{Benchmark Task.} 
The task of the benchmark is to compare the similarity of tactics between the team in real world football video fragment and the one in virtual football simulation platform. We only evaluate the trajectory of the ball as it closely reflects offensive intent and team rhythm while being more sensitive to spatial/directional biases. In contrast, we do not score individual player trajectories: role changes, occlusions, and identity ambiguities in the game introduce high-variance matching noise, which bias the evaluation toward individual alignment over overall tactical style.
During testing, we only provide the \emph{first frame} context (player and ball positions) and let the model infer about the subsequent process, using the testset to evaluate the inference results. Throughout training, no additional task-specific constraints or manually defined rules are imposed beyond uniform preprocessing and observation Windows. 
\section{Experiments}
\subsection{Baseline Models}
\begin{table*}[!t]
	\centering
	\caption{
		\textbf{Performance comparison under different grid resolutions.}
		We evaluate imitation accuracy across four representative models under multiple spatial grid configurations. 
		Spatial occupancy and movement vector similarity (\%) are reported. 
		Moderate grid sizes yield the most balanced and stable results, while excessively coarse or fine grids lead to degraded spatial alignment and tactical coherence.
	}
	\label{tab:main_results}
	\renewcommand{\arraystretch}{1.2} 
	\small 
	\begin{tabular*}{\textwidth}{@{\extracolsep{\fill}}lccccccccccc}
		\toprule
		\multirow{2}{*}{\textbf{Grids (L × W)}} &
		\multirow{2}{*}{\textbf{Proportion (L/W)}} &
		\multirow{2}{*}{\textbf{Method}} &
		\multicolumn{3}{c}{\textbf{Average(3.0s)}} &
		\multicolumn{3}{c}{\textbf{Average(5.0s)}} &
		\multicolumn{3}{c}{\textbf{Average(10.0s)}} \\
		\cmidrule(lr){4-6} \cmidrule(lr){7-9} \cmidrule(lr){10-12}
		& & & Score & $S_t$ & $S_v$ & Score & $S_t$ & $S_v$ & Score & $S_t$ & $S_v$ \\
		\midrule
		\multirow{5}{*}{60 ($10 \times 6$)} & \multirow{5}{*}{0.9259} 
		& BC~\cite{pomerleau1988alvinn}        & 37.05 & 32.86 & 41.24 & 21.19 & 19.84 & 22.54 & 11.04 & 10.73 & 11.35 \\
		& & CMIL~\cite{le2017coordinated}      & 46.06 & 41.34 & \textbf{50.78} & 31.86 & 28.43 & 35.29 & 27.57 & \textbf{28.48} & 26.66 \\
		& & IRL~\cite{rahimian2021inferring}   & 34.18 & 30.12 & 38.24 & 22.16 & 20.57 & 23.74 & 13.81 & 12.13 & 15.48 \\
		& & CoDAIL~\cite{Liu2020Multi-Agent}   & \textbf{46.63} & 42.92 & 50.34 & \textbf{40.21} & \textbf{38.45} & \textbf{41.97} & \textbf{33.00} & \textbf{29.44} & \textbf{36.56} \\
		& & DRAIL~\cite{lai2024diffusion}      & 45.41 & \textbf{43.45} & 47.37 & 29.61 & 33.44 & 25.78 & 28.65 & 27.55 & 29.74 \\
		\midrule
		\multirow{5}{*}{150 ($15 \times 10$)} & \multirow{5}{*}{1.0294} 
		& BC~\cite{pomerleau1988alvinn}        & 37.86 & 28.57 & 47.14 & 18.78 & 15.94 & 21.62 &  8.63 &  8.01 &  9.25 \\
		& & CMIL~\cite{le2017coordinated}      & 42.98 & 40.22 & 45.73 & 33.09 & 31.35 & 34.82 & 26.35 & \textbf{28.47} & 24.22 \\
		& & IRL~\cite{rahimian2021inferring}   & 32.53 & 28.34 & 36.72 & 19.93 & 18.34 & 21.52 & 11.61 & 10.65 & 12.56 \\
		& & CoDAIL~\cite{Liu2020Multi-Agent}   & \textbf{50.89} & \textbf{48.56} & \textbf{53.22} & \textbf{39.39} & \textbf{38.56} & \textbf{40.22} & \textbf{28.37} & 20.01 & \textbf{36.72} \\
		& & DRAIL~\cite{lai2024diffusion}      & 41.72 & 39.88 & 43.56 & 26.72 & 29.88 & 23.56 & 26.97 & 26.49 & 27.44 \\
		\midrule
		\multirow{5}{*}{240 ($20 \times 12$)} & \multirow{5}{*}{0.9259} 
		& BC~\cite{pomerleau1988alvinn}        & 24.90 & 18.18 & 31.62 &  5.98 &  4.33 &  7.62 &  8.26 &  7.11 &  9.41 \\
		& & CMIL~\cite{le2017coordinated}      & \textbf{47.87} & \textbf{45.33} & \textbf{50.41} & 29.81 & 29.81 & 29.81 & 20.11 & 20.41 & 19.81 \\
		& & IRL~\cite{rahimian2021inferring}   & 28.33 & 24.21 & 32.45 & 16.12 & 14.79 & 17.45 &  9.25 &  8.49 & 10.01 \\
		& & CoDAIL~\cite{Liu2020Multi-Agent}   & 43.22 & 42.88 & 43.56 & \textbf{31.40} & \textbf{30.34} & \textbf{32.45} & 17.10 & 16.32 & 17.88 \\
		& & DRAIL~\cite{lai2024diffusion}      & 31.73 & 31.34 & 32.11 & 26.61 & 30.24 & 22.97 & \textbf{22.64} & \textbf{24.05} & \textbf{21.22} \\
		\midrule
		\multirow{5}{*}{600 ($30 \times 20$)} & \multirow{5}{*}{1.0294} 
		& BC~\cite{pomerleau1988alvinn}        & 19.12 & 15.78 & 22.45 & 18.98 & 13.33 & 24.62 &  9.24 &  6.67 & 11.81 \\
		& & CMIL~\cite{le2017coordinated}      & 35.54 & \textbf{36.33} & 34.75 & \textbf{25.70} & \textbf{24.65} & \textbf{26.74} & 15.65 & 14.56 & 16.74 \\
		& & IRL~\cite{rahimian2021inferring}   & 23.05 & 18.44 & 27.65 & 14.11 & 12.98 & 15.23 &  8.14 &  7.11 &  9.17 \\
		& & CoDAIL~\cite{Liu2020Multi-Agent}   & \textbf{37.12} & 34.78 & \textbf{39.45} & 20.90 & 20.34 & 21.45 & 14.12 & 10.67 & 17.56 \\
		& & DRAIL~\cite{lai2024diffusion}      & 30.19 & 28.45 & 31.92 & 20.69 & 23.45 & 17.92 & \textbf{20.69} & \textbf{23.45} & \textbf{17.92} \\
		\midrule
		\multirow{5}{*}{1768 ($105 \times 68$)} & \multirow{5}{*}{1.0005} 
		& BC~\cite{pomerleau1988alvinn}        & 10.85 & 10.94 & 10.75 &  6.50 &  4.65 &  8.34 &  5.03 &  3.55 &  6.51 \\
		& & CMIL~\cite{le2017coordinated}      & 23.12 & 22.44 & 23.79 & 17.12 & 15.74 & 18.49 &  8.26 &  7.11 &  9.41 \\
		& & IRL~\cite{rahimian2021inferring}   & 17.31 & 13.27 & 21.34 & 10.61 &  8.99 & 12.22 &  6.43 &  5.64 &  7.21 \\
		& & CoDAIL~\cite{Liu2020Multi-Agent}   & \textbf{27.10} & \textbf{26.32} & \textbf{27.88} & 13.62 & 11.58 & 15.66 &  6.45 &  5.34 &  7.56 \\
		& & DRAIL~\cite{lai2024diffusion}      & 22.14 & 21.05 & 23.22 & \textbf{19.14} & \textbf{17.05} & \textbf{21.22} & \textbf{10.64} & \textbf{10.05} & \textbf{11.22} \\
		\bottomrule
	\end{tabular*}
\end{table*}

To validate the effectiveness of TacSIm as a benchmark for tactical imitation, 
we evaluate four representative imitation-learning baselines that capture different perspectives of coordinated multi-agent behavior. 
All models are trained and tested under identical observation–prediction settings and dataset splits described in Section~3.2, ensuring a fair comparison.

\textbf{Behavior Cloning (BC)}~\cite{pomerleau1988alvinn}. 
BC serves as the simplest form of imitation learning, directly mapping observed states to expert actions via supervised learning. 
It establishes a lower bound for tactical imitation, providing a reference for how well purely supervised policies can reproduce collective player movements.

\textbf{Coordinated Multi-Agent Imitation Learning (CMIL)}~\cite{le2017coordinated}. 
CMIL extends classical imitation learning to multi-agent settings by introducing coordination mechanisms among agents. 
Through shared latent representations, it enables agents to implicitly model inter-player dependencies, improving group-level consistency compared with independent BC policies.

\textbf{Inverse Reinforcement Learning (IRL)}~\cite{rahimian2021inferring}. 
IRL aims to learn the underlying reward function that drives expert behavior by observing demonstrations. It allows the agent to infer the expert's implicit goals and strategies, improving the agent's ability to generalize across different tactical situations. 
IRL models have the advantage of learning from demonstrated behavior without requiring explicit action labels, making them a powerful tool for modeling complex, coordinated team tactics.

\textbf{Decentralized Adversarial Imitation Learning algorithm with Correlated policies (CoDAIL)}~\cite{Liu2020Multi-Agent}. 
CoDAIL adopts an adversarial learning framework where a discriminator distinguishes real and generated team trajectories. 
It combines diffusion-based noise modeling with coordination priors to stabilize multi-agent policy learning and generate more coherent tactical behaviors across players.

\textbf{Diffusion Rewards Adversarial Imitation Learning (DRAIL)}~\cite{lai2024diffusion}. 
DRAIL leverages diffusion processes for imitation learning, progressively denoising latent tactical representations toward realistic trajectories. 
Compared to CoDAIL, it focuses on the generative smoothness and long-horizon consistency of player coordination, enabling stable reconstruction of team-level tactical dynamics.

\subsection{Implementation Details}
To unify the training and testing environments, we employ a multi-window length approach for processing video clips. During training, the model receives randomly sampled context lengths $L \in \{1,10,25,50\}$ (including $L = 1$ to match the testing environment) and optimizes predictions for the subsequent 25 frames. The loss value is averaged across all windows. We incorporate short-term closed-loop inference, using a small number of predicted steps as input feedback and supervising against ground truth—to mitigate exposure bias.
Full hyperparameters are listed in the Supplement.

Experiments are conducted on an NVIDIA 3090 24G GPU with PyTorch 2.3. Each experiment is repeated three times with different random seeds, and we report the mean and standard deviation of all metrics.

\subsection{Analysis of results} 
\begin{figure*}[t]
	\centering
	\includegraphics[width=\linewidth]{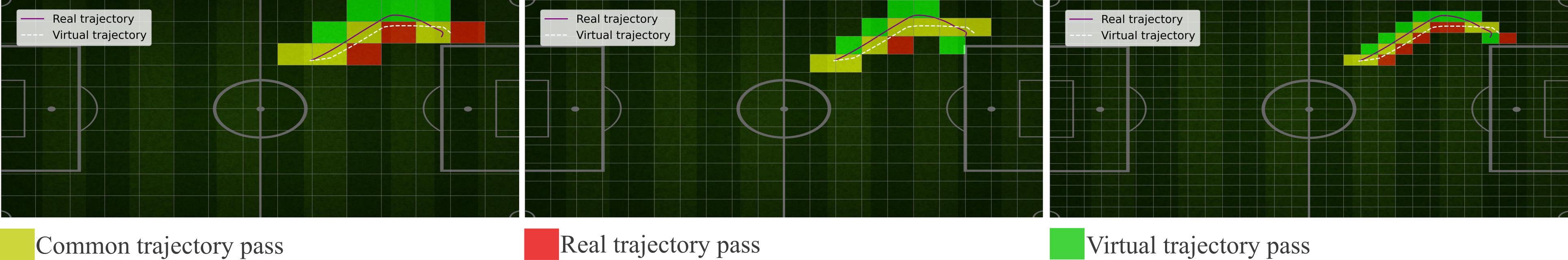}
	\caption{
		Visualization of tactical imitation.This figure shows the comparison between Ground Truth (purple solid line) and Inference (white dashed line) for player trajectory predictions in different tactical contexts. The colorful blocks on the field represent the spatial grid, showing the distribution of the player's movement across different zones during the match. From left to right, the grid resolutions are $15 \times 10$, $20 \times 12$, and $30 \times 20$, illustrating how varying granularity affects spatial coverage and prediction consistency.}
	\label{fig:ruslet}
\end{figure*}

Based on the experimental results in Table~\ref{tab:main_results}, we conduct an analysis combining grid scale (coarse-fine) and prediction time (short-long). Crossing these axes reveal four distinct patterns and method preferences. Figure~\ref{fig:ruslet} shows a visual representation of the ball trajectory at partial grid sizes. For more detailed information, please refer to the supplementary materials.

Prediction duration is the primary factor causing systematic degradation in model performance. Evaluation metrics show a significant decline across all models from short-term (3.0s) to medium-term (5.0s) and long-term (10.0s) predictions. This phenomenon reveals an inherent challenge in trajectory prediction tasks: prediction uncertainty increases as the time horizon extends. In the short-term phase, models primarily rely on initial motion states for local extrapolation, making the task relatively straightforward. However, when transitioning to medium- and long-term predictions, models must leap from low-level “motion state mimicry” to higher-level “tactical intent deduction.” This requires understanding dynamic strategies such as player coordination and offensive-defensive transitions. The widespread performance degradation, particularly the steep decline of traditional behavior cloning (BC) methods, indicates that most models still face significant bottlenecks in capturing long-range spatio-temporal dependencies and strategic reasoning.
There exists an optimal “golden range” for spatial discretization grid resolution. Experimental results indicate that medium-sized grid configurations (15×10 with 150 cells and 20×12 with 240 cells) provide the optimal performance balance in the vast majority of cases. Coarse-grained grids (60 cells) inherently lose information, failing to capture intricate positioning and tactical details, resulting in overly simplistic strategies learned by the model. Conversely, excessively fine grids (1768) fall into the “curse of dimensionality” dilemma. The extreme sparsity of the state space makes it difficult for the model to inductively learn effective patterns from limited data. Simultaneously, it may force the model to overemphasize insignificant displacement noise lacking tactical significance, thereby compromising its generalization ability. Therefore, a moderate discretization scale achieves a favorable balance between preserving critical tactical information and maintaining model learnability.

Through cross-analysis of time and grid size, a profound interaction between temporal and spatial dimensions has been identified. This interaction can be applied across various application scenarios. (1) “Short-term/Fine-grained” task scenarios: When task objectives involve micro-simulations for individual player technique analysis, one-on-one dribbling, or pass-and-shoot decision-making, short-term predictions (3.0s) paired with moderately fine grids yield optimal results. Short-term-Fine" Task Scenarios: When task objectives involve micro-simulations for individual player skill analysis, one-on-one dribbling, or pass-shoot decision-making, short-term predictions (3.0s) paired with moderately fine grids (150/240) form the optimal configuration. In this scenario, the model fully leverages detailed spatial information to accurately capture subtle technical actions like acceleration, direction changes, and ball contact, meeting micro-analysis requirements. (2) “Long-term - Macro” Task Scenarios: When task objectives shift toward macro tactical simulations, formation evolution analysis, or overall offensive pattern modeling, long-term predictions (10.0s) paired with medium-coarse grids (150/60) demonstrate unique value. A larger grid size guides the model to predict the “tactical zones” players will ultimately occupy, rather than pinpointing exact coordinates. This approach mirrors the strategic mindset of coaches when deploying formations.

\subsection{Discussion}
The experimental results of this study reveal the unique advantages of different imitation learning models in temporal and spatial dimensions, providing clear guidance for future research directions. Specifically, the outstanding performance demonstrated by the CoDAIL model in medium-to-short-term predictions and at medium grid resolutions highlights the effectiveness of its explicit modeling of multi-agent coordination mechanisms. This indicates its significant potential in tasks requiring high-fidelity, detailed tactical action simulation. Diffusion models like DRAIL demonstrate relative robustness in long-term forecasting, particularly regarding the spatial occupancy ($S_t$) metric, suggesting their potential for macro-strategic simulation and predicting long-duration formation dynamics. Future research should not pursue a “universal” solution but instead focus on developing multi-scale, hierarchical frameworks that integrate the strengths of both approaches ultimately achieving a soccer trajectory imitation learning system that combines tactical precision with strategic depth.

\section{Conclusion}
In this paper, we introduce TacSIm, the first large-scale dataset and benchmark for multi-agent tactical imitation for football, which combines real Premier League broadcast trajectory data with a structured tactical phase taxonomy. This benchmark bridges the gap between real-world match dynamics and virtual simulation, offering an adaptive-grid evaluation and phase-aware protocol for assessing coordinated team behaviors.   Experiments  demonstrate that TacSIm effectively distinguishes between models that merely extrapolate motion and those that capture collective tactical organization.  
We hope that TacSIm will serve as a foundation for advancing tactical modeling, with future expansions incorporating richer semantic layers and multimodal inputs to enhance multi-agent imitation and sports analytics.

{
	\small
	\bibliographystyle{ieeenat_fullname}
	\bibliography{main}
}


\end{document}